\documentclass[journal]{IEEEtran}

\usepackage{cite}
\usepackage{graphicx}
\usepackage{textcomp}
\usepackage{wrapfig}
\usepackage[utf8]{inputenc}
\usepackage{authblk}
\usepackage{graphicx}
\usepackage{cite}
\usepackage{indentfirst}
\usepackage[justification=centering]{caption}
\usepackage{amsmath}
\usepackage{lscape}
\usepackage{gensymb}
\usepackage{algorithm}
\usepackage{algpseudocode}
\usepackage{mathtools}
\usepackage{amsfonts}
\usepackage{comment}
\usepackage{bm}
\usepackage{multirow}
\usepackage{marvosym}
\usepackage[dvipsnames]{xcolor}
\algnewcommand{\Inputs}[1]{%
  \State \textbf{Inputs:}
  \Statex \hspace*{\algorithmicindent}\parbox[t]{.8\linewidth}{\raggedright #1}
}
\algnewcommand{\Initialize}[1]{%
  \State \textbf{Initialize:}
  \Statex \hspace*{\algorithmicindent}\parbox[t]{.8\linewidth}{\raggedright #1}
}
\usepackage{hyperref}
\usepackage[dvipsnames]{xcolor}

\usepackage{graphicx}
\usepackage{array}
\usepackage{makecell}

\graphicspath{ {Images/} }

\begin{document}

\title{Multi-Modal Attention Networks for Enhanced Segmentation and Depth Estimation of Subsurface Defects in Pulse Thermography}

\author{Mohammed Salah,
        Naoufel Werghi,
        Davor Svetinovic,
        and~Yusra~Abdulrahman (Member, IEEE)
\thanks{The work was supported by Khalifa University of Science and Technology.}
\thanks{M. Salah and Y. Abdulrahman are with the Department of Aerospace Engineering, Khalifa University, Abu Dhabi, UAE.

N. Werghi and D. Svetinovic are with the Department of Computer Science, Khalifa University, Abu Dhabi, UAE.

Yusra Abdulrahman is the corresponding author (email: yusra.abdulrahman@ku.ac.ae)}}

\maketitle

\begin{abstract}
AI-driven pulse thermography (PT) has become a crucial tool in non-destructive testing (NDT), enabling automatic detection of hidden anomalies in various industrial components. Current state-of-the-art techniques feed segmentation and depth estimation networks compressed PT sequences using either Principal Component Analysis (PCA) or Thermographic Signal Reconstruction (TSR). However, treating these two modalities independently constrains the performance of PT inspection models as these representations possess complementary semantic features. To address this limitation, this work proposes PT-Fusion, a multi-modal attention-based fusion network that fuses both PCA and TSR modalities for defect segmentation and depth estimation of subsurface defects in PT setups. PT-Fusion introduces novel feature fusion modules, Encoder Attention Fusion Gate (EAFG) and Attention Enhanced Decoding Block (AEDB), to fuse PCA and TSR features for enhanced segmentation and depth estimation of subsurface defects. In addition, a novel data augmentation technique is proposed based on random data sampling from thermographic sequences to alleviate the scarcity of PT datasets. The proposed method is benchmarked against state-of-the-art PT inspection models, including U-Net, attention U-Net, and 3D-CNN on the Université Laval IRT-PVC dataset. The results demonstrate that PT-Fusion outperforms the aforementioned models in defect segmentation and depth estimation accuracies with a margin of 10\%. 
\end{abstract}

\IEEEpeerreviewmaketitle

\section*{Code and Multimedia Material}

\begin{center}
Video: \\ \href{https://drive.google.com/file/d/11TH-YWYLilyIkBk-K5XXn8Ox3msOE0V2/view?usp=drive_link}{Video Link} \\
Code: \\
\url{https://github.com/mohammedsalah98/pt_fusion}

\end{center}

\section{Introduction} \label{sec:introduction}
\IEEEPARstart{O}{ver} the recent years, the growing demand for quality control drove the emergence of advanced inspection processes in various production lines, such as the aerospace \cite{ndt_review, 3d_cnn} automotive \cite{automotive_1, automotive_2}, and construction \cite{construction_ndt} industries. This paradigm shift led to the adoption of Non-Destructive Testing (NDT) techniques to detect subsurface defects and assess the structural integrity of industrial components. As a result, a large spectrum of NDT techniques have been developed to detect subsurface defects, such as radiography, ultrasonic testing, and infrared thermography (IRT) \cite{irt_survey}.

IRT has garnered significant attention due to its simplicity, low cost, and non-contact nature. These advantages have demonstrated the potential of IRT for inspection of aircraft parts \cite{yusra_3, cfrp_deep, 3d_cnn, yusra_4}, construction pipelines \cite{construction_ndt, construction_uav, uav_solar, drone_sites}, and artworks \cite{autoencoder, artwork}. Among IRT inspection techniques, pulse thermography (PT) is valued for its efficiency in inspecting large areas and its capability to rapidly identify subsurface defects \cite{yusra_taguchi}. Meanwhile, recent advancements in artificial intelligence (AI) have been adopted in PT setups to enhance inspection accuracies. For instance, AI-driven PT inspection algorithms have been proposed for defect classification \cite{flexible_framework, detection}, segmentation \cite{attention_unet, pt_dataset}, and depth estimation \cite{irt_depth, 3d_cnn}.

State-of-the-art PT inspection methods feed AI models compressed thermographic sequences using principal component analysis (PCA) \cite{pct} or thermographic signal reconstruction (TSR) \cite{tsr} for defect characterization. While these methods demonstrated considerable potential, treating PCA and TSR as two independent modalities limits the performance of PT inspection models. For instance, PCA tends to enhance defect visibility while TSR models the temporal evolution of the pixel responses. To address this limitation, this work introduces PT-Fusion, a multi-modal attention-based fusion network that fuses both PCA and TSR complementary representations for segmentation and depth estimation of subsurface defects in PT setups. The contributions of the paper are the following:

\begin{enumerate}
    \item PT-Fusion network architecture is proposed for defect segmentation and depth estimation in pulse thermography setups.
    \item Novel feature fusion modules, Encoder Attention Fusion Gate (EAFG) and Attention Enhanced Decoding Block (AEDB) are proposed to fuse PCA and TSR representations for enhanced segmentation and depth estimation of subsurface defects.
    \item A novel data augmentation technique is introduced, leveraging random sampling from thermographic sequences to overcome the burden of complex PT data collection setups.
    \item The proposed work is tested on the IRT-PVC dataset and benchmarked against state-of-the-art PT inspection models. The results demonstrate that PT-Fusion outperforms U-Net, attention U-Net, and 3D-CNN architectures in segmentation and depth estimation accuracies.
\end{enumerate}

\subsection{Related Work} \label{subsection:related_work}
Pulse thermography (PT) has established itself as a key NDT technique for the detection of subsurface defects in industrial components. The recent advancements in artificial intelligence (AI) have significantly enhanced PT methodologies, enabling automated defect characterization. As a result, extensive investigations have emerged on AI-driven PT leveraging learning-based methods, with various architectures proposed for defect detection, segmentation, and depth estimation. For instance, a broad spectrum of learning-based defect detection models have been developed demonstrating the potential of PT. Muller et al. \cite{cnnlstm} introduced a ConvLSTM network for the detection of flat defects in forged steel parts. Consequently, Ankel et al. \cite{tomography_cnn} extended convolutional neural networks (CNNs) for elliptical defect detection. Other state-of-the-art detection networks \cite{tim_detection} have been developed for defect detection, such as YOLOv5 \cite{irt_depth} and Faster R-CNN \cite{flexible_framework}. Since these methods are purely convolutional, the performance of CNN-based architectures is limited by the network kernels receptive fields. Thus, an attention-based defect detection network was introduced to simultaneously learn local and global thermal features, circumventing the limited receptive fields of CNNs \cite{attention_spatiotemporal}. Other methods leverage generative models for data augmentation and progressive training for improved learning of defect detection networks \cite{gan, diffusion}.

Similarly, a wide stream of defect segmentation and depth estimation networks have been proposed.  Fang et al. \cite{unet_study} performed an experimental study on state-of-the-art network architectures, such as U-Net and ResNet, for subsurface defect segmentation. Zhou et al. \cite{attention_unet} introduced the attention U-Net for enhanced defect segmentation compared to its traditional U-Net counterpart. On the other hand, Morelli et al. \cite{tim_2} developed a residual convolutional network for heat-invariant defect segmentation in lock-in thermography. The aforementioned networks are inherently 2D with minimal consideration of the significant temporal information embedded in thermographic sequences. Thus, Dong et al. \cite{3d_cnn} proposed a 3D-CNN accounting for temporal information embedded in raw thermal sequences for segmentation and depth estimation of CFRP subsurface defects using lock-in thermography. However, all these networks are completely convolutional, and the limited receptive field of CNNs limits their performances. Consequently, several works leverage attention-based architectures for segmentation models to learn global spatial and temporal features. Schmid et al. \cite{schmid} and Xie et al. \cite{xie} proposed temporal attention mechanisms for improving learned features from temporal sequences. In addition, attention-based PHM-IRNET network architecture was proposed as a self-supervised learning framework for segmentation in industrial thermographic inspection \cite{phm}. 

In all of the aforementioned methods, input to the segmentation networks is single image modalities such as raw thermal sequences, pulse-phase thermography (PPT) \cite{ppt_1, ppt_2}, PCA \cite{pct, pca_2, pca_3}, or TSR \cite{tsr, tsr_2, tsr_3}. Other approaches rely on learning-based data enhancement methods, such as 1D-CNN \cite{1d_cnn} and deep autoencoders \cite{autoencoder}, as single inputs to learning-based methods. However, these representations possess semantic features that are rather complimentary. For instance, PCA tends to enhance defect visibility while TSR models the temporal evolution of the pixel responses. Unlike state-of-the-art methods, this work proposes learning these complimentary features in an end-to-end network, PT-Fusion, for subsurface defect segmentation and depth estimation. PT-Fusion passes PCA and TSR modalities into a CNN head, and the extracted feature maps are fed to novel fusion blocks, EAFG and AEDB, to dynamically learn spatial and temporal features for enhanced segmentation and depth estimation.


The rest of the article is structured as follows. Section \ref{section:preliminaries} provides necessary preliminaries for PT-Fusion methodology. Section \ref{section:methodology} outlines the proposed network architecture and data augmentation strategy. Section \ref{section:exp} presents the experimental validations of the proposed fusion network tested on the IRT-PVC dataset. Finally, section \ref{section:conc} presents conclusions, findings, and future aspects of the proposed work.
 
\section{Preliminaries} \label{section:preliminaries}
\subsection{Pulse Thermography} \label{subsection:pt}
\begin{figure}[t]
\center
\includegraphics[scale=0.45]{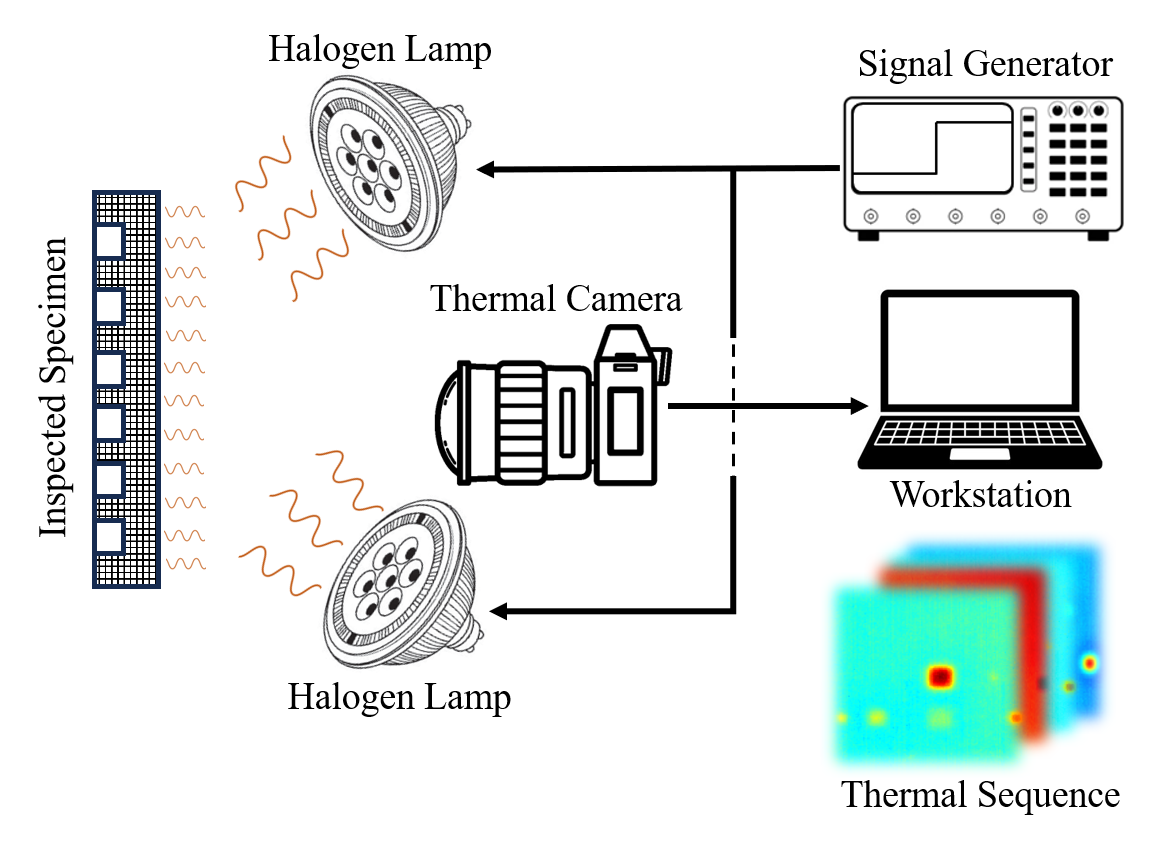}
\caption{PT setup involving flash lamps providing heat pulse to a specimen. Defective areas trap heat flow generating abnormal temperature distributions.}
\label{fig:pt_setup}
\end{figure}

A typical PT setup is shown in Fig. \ref{fig:pt_setup}. PT involves a flash lamp providing a heat pulse into a target specimen \cite{ndt_survey}. If the specimen is sound, all pixels approximately generate the same thermal profile. In the presence of defects, heat gets trapped, generating an abnormal thermal distribution that is captured over time by the thermal camera. Consequently, the acquired thermal images are pre-processed using the methods discussed in section \ref{subsection:data_collection} and are utilized as inputs to PT-Fusion.

\subsection{IRT-PVC Dataset and Data Preprocessing} \label{subsection:data_collection}
Several public datasets are available following the general PT setup discussed in section \ref{subsection:pt}. In this work, PT-Fusion is trained and tested on the IRT-PVC dataset \cite{irt_depth}. The dataset comprises 38 inspection sequences of 3D-printed PVC specimens with back-drilled holes at different depths ranging between 0 and 2.5 mm. Each inspection sequence $\mathbf{S}_{i} = \{ I_{k} \}_{1}^{N_{t}}$ is a 3D matrix of shape $(N_{t}, N_{y}, N_{x})$ with $I_{k}$ thermograms, where $k = 1, 2, \dots, N_{t}$ is the image timestamp, $i$ represents the sample index, $N_{y}$ is the image height, and $N_{x}$ is its width. $\mathbf{S}_{i}$ is reshaped to $(N_{t}, N_{y} \times N_{x})$ by a raster-like operation and standardized by

\begin{equation}
    \mathbf{\hat{S}}_{i} = \frac{\mathbf{S}_{i} - \mu_{k}}{\sigma_{k}},
\end{equation}

\noindent where,
\begin{equation}
    \mu_{k} = \frac{1}{N_{t}} \sum_{k=1}^{N_{t}} S^{(k)},
\end{equation}

\begin{equation}
    \sigma_{k}^{2} = \frac{1}{N_{t} - 1} \sum_{k=1}^{N_{t}}(S^{(k)} - \mu_{k})^{2},
\end{equation}

\begin{figure}[b]
\center
\includegraphics[scale=0.33]{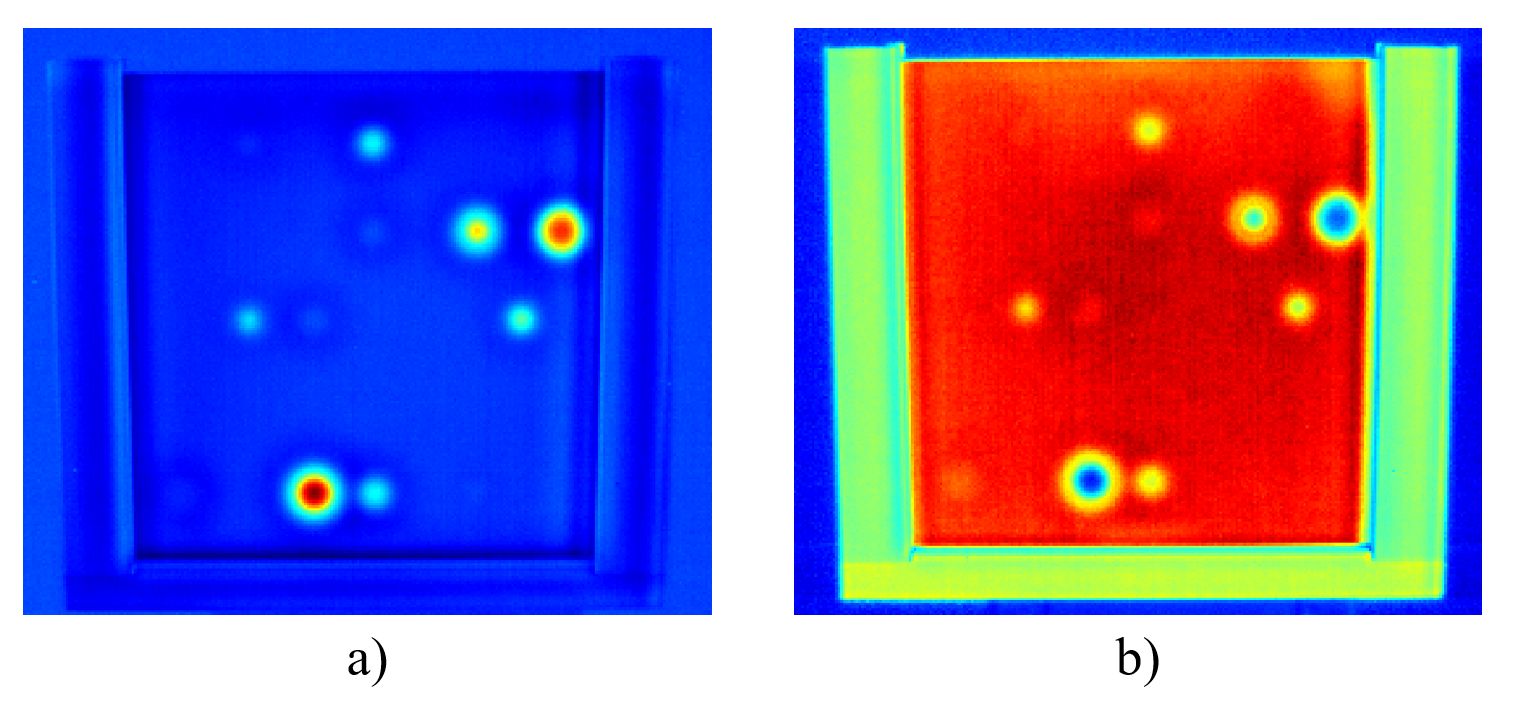}
\caption{Generated thermographic representations, a) PCA and b) TSR, for sequence \textit{Z\_013} in IRT-PVC dataset.}
\label{fig:pca_tsr}
\end{figure}

\noindent and $\mathbf{\hat{S}}_{i} = [S^{(1)}, S^{(2)}, \dots, S^{(N_{t})}]$ is the standardized pixel-wise thermal response. Developing PT inspection algorithms that process the entire thermal sequence introduces significant computational complexities. Instead, data compression techniques such as PCA and TSR are employed to generate PCA and TSR images. These thermographic representations offer the advantage of reducing data dimensionality while preserving the most important features for defect detection. PCA enhances defect clarity, while TSR incorporates the temporal evolution of the pixel responses.

To generate PCA images, singular value decomposition (SVD) is applied by

\begin{equation}
    \mathbf{\hat{S}}_{i} = \mathcal{U}\Gamma V^{T},
\end{equation}

\noindent where $\mathcal{U}$ comprises empirical orthogonal functions representing spatial differences within the input pixel responses, $\Gamma$ is a diagonal matrix comprising singular eigenvalues in descending order, while matrix $V$ contains their corresponding principal components. Consequently, the principal component images $P_{k}$ are computed by

\begin{equation}
    P_{k} = \mathbf{\hat{S}}_{i}v_{k},
\end{equation}

\begin{figure*}[!h]
    \centering
    \includegraphics[keepaspectratio=true,scale=0.475, width=\linewidth]{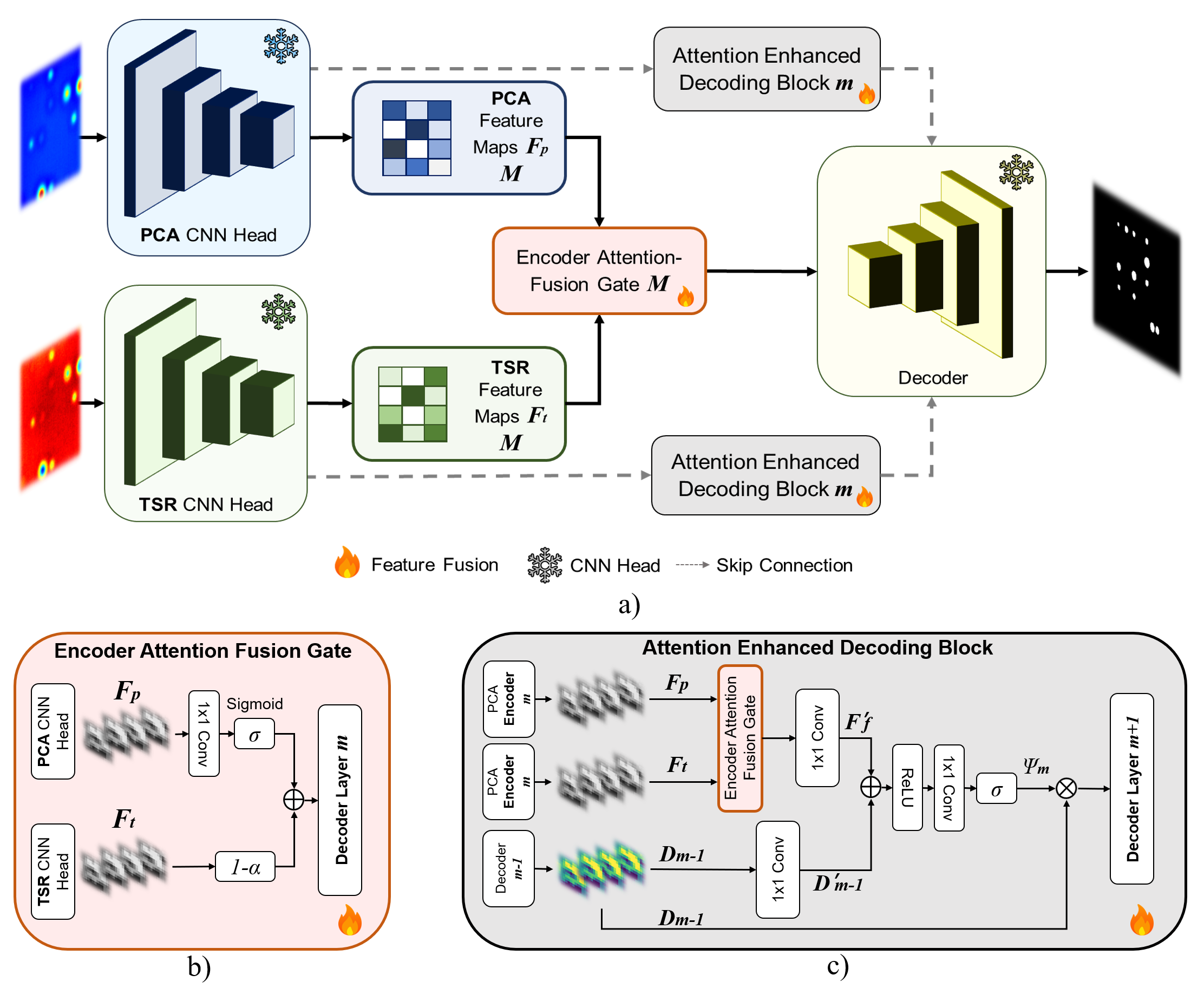}
    \caption{a) PT-Fusion network architecture. PCA and TSR images are fed to a shallow CNN head. b) The Feature fusion module, Encoder Attention Fusion Gate (EAFG), is placed at the output of the encoders, and c) the Attention Enhanced Decoding Block (EADB) is added to adaptively fuse encoded features with the decoders.}
    \label{fig:network}
\end{figure*}

\noindent where $v_{k}$ is the $k^{th}$ eigenvector in the matrix $V$. It is worth mentioning that only the first $J=$ 5-10 principal component images are utilized for inspection \cite{pct}, where in our study $J=10$ is selected to generate the PCA image tensor $\mathbf{P}_{i} \in \mathbb{R}^{J \times N_{y} \times N_{x}}$, where $\mathbf{P}_{i} = [P_{1}, P_{2}, \dots, P_{J}]$.

On the other hand, TSR image generation involves logarithmic polynomial fitting on the pixel's thermal response with an n-degree polynomial as

\begin{equation}
\begin{split}
    \ln{(\Delta T)} &= a_{0} + a_{1}\ln{(t)} + a_{2}\ln{(t)}^{2} \\
    &+ a_{3}\ln{(t)}^{3} + a_{4}\ln{(t)}^{4} + a_{5}\ln{(t)}^{5},
\end{split}
\end{equation}

\noindent where $a_{n}$ are the coefficients of the $n$-degree polynomial, $\Delta T$ is the temperature increase, and $t$ is time. 4-5 degree polynomials are sufficient to model the pixel responses. In this work, an $n=5$ degree polynomial is employed. The resulting polynomial coefficients are used as the values for the TSR image channels, forming the TSR image tensor $\mathbf{T}_{i} \in \mathbb{R}^{n \times N_{y} \times N_{x}}$ and $\mathbf{T}_{i} =[T_{1}, T_{2}, \dots, T_{n}]$. Collectively, each sequence generates a 10 channel PCA image and 5 channel TSR image, see Fig. \ref{fig:pca_tsr}, and are utilized as inputs to the proposed PT-Fusion network.

\section{Methodology} \label{section:methodology}

\subsection{PT-Fusion Architecture} \label{subsection:architecture}
PT-Fusion integrates the complementary features of PCA, $\mathbf{P}_{i}$, and TSR, $\mathbf{T}_{i}$. Generally, the direct approach for fusing both modalities is by either concatenating them at the input or concatenating their feature maps in the intermediate layers. However, the network, in such cases, does not inherently learn the weighted contribution of these modalities, limiting its performance. Instead, PT-Fusion adaptively fuses both representations using the proposed attention-based fusion modules, EAFG and AEDB.

\begin{figure}[t]
\center
\includegraphics[scale=0.3]{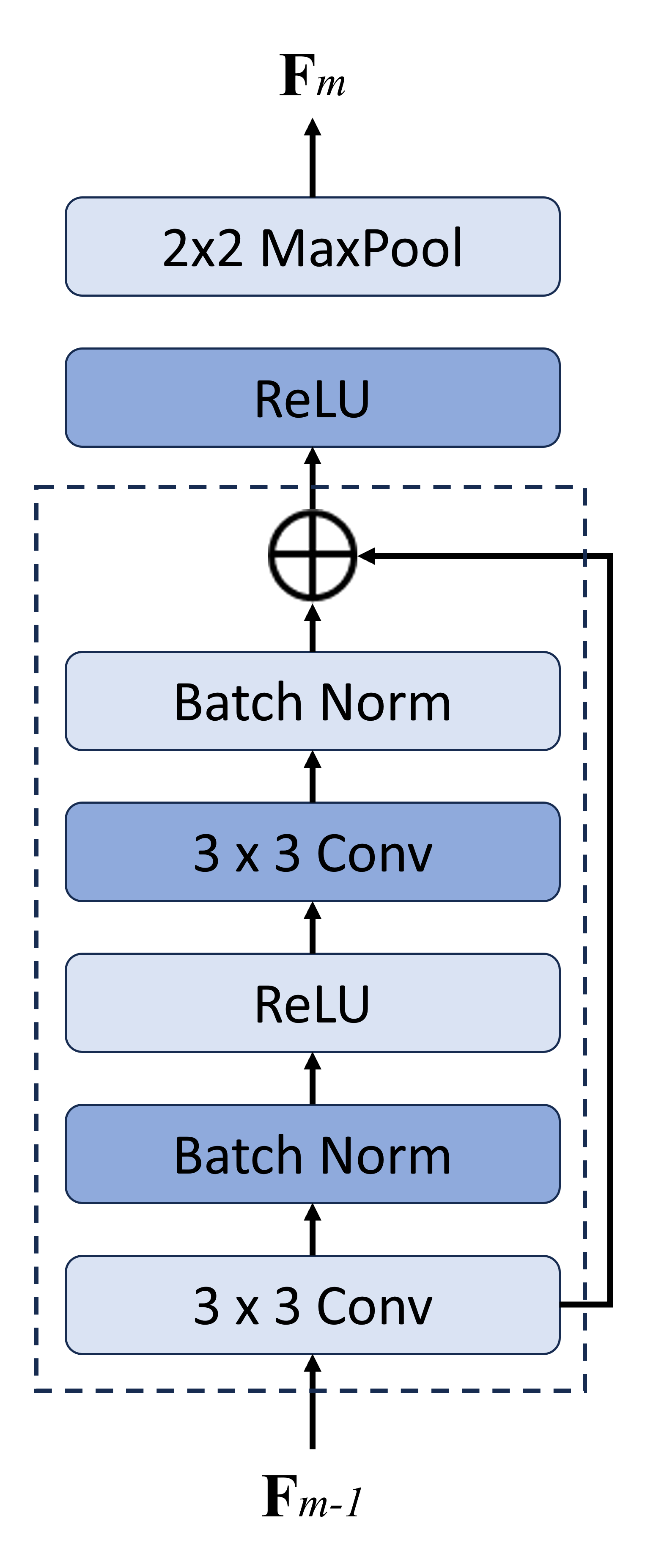}
\caption{Encoder layer $m$ of the PT-Fusion CNN heads involving a residual convolution block and downsampling max pooling for feature extraction.}
\label{fig:res_conv}
\end{figure}

PT-Fusion, illustrated in Fig. \ref{fig:network}, follows a U-Net encoder-decoder architecture \cite{unet} and involves passing $\mathbf{P}_{i}$ and $\mathbf{T}_{i}$ through two parallel CNN heads to extract feature maps from both modalities. Each CNN head consists of $m$ encoder layers, where $m = 1, 2, \dots, M$. Each layer, see Fig. \ref{fig:res_conv}, feeds its input through a learnable residual convolution with a downsampling max pooling operation as

\begin{equation}
    \mathbf{F}_{m}^{p} = \text{ResidualConv}\big(\text{MaxPool}(\mathbf{F}_{m-1}^{p})\big)
\end{equation}

\begin{equation}
    \mathbf{F}_{m}^{t} = \text{ResidualConv}\big(\text{MaxPool}(\mathbf{F}_{m-1}^{t})\big)
\end{equation}

\noindent where $\mathbf{F}_{m}^{p}$ and $\mathbf{F}_{m}^{t}$ are the m-th encoder layer outputs of the PCA and TSR CNN heads. Similarly, $\mathbf{F}_{m-1}^{p}$ and $\mathbf{F}_{m-1}^{t}$ are the layer inputs received from the $m-1$ encoder layer.

After encoding the input from the PCA and TSR representations to the feature maps generated from the last encoder layers $\mathbf{F}_{M}^{p}$ and $\mathbf{F}_{M}^{t}$, the features maps are fused in EAFG using their weighted combination to exploit the complementary features. EAFG is governed by learnable attention weights $\boldsymbol{\alpha}_{M}$, derived from the PCA features as

\begin{equation}
    \boldsymbol{\alpha}_{M} = \sigma\big(\mathbf{W}^{1\times 1}_{M} (\mathbf{F}_{M}^{p}) \big),
\end{equation}

\noindent where $\mathbf{W}^{1\times 1}_{M}$ represents a $1 \times 1$ convolution, while $\sigma$ is the sigmoid activation. Consequently, the fused feature map $\mathbf{F}^{f}_{M}$ is computed by

\begin{equation}
    \mathbf{F}^{f}_{M} = \boldsymbol{\alpha}_{M} \otimes \mathbf{F}_{M}^{p} + (1 - \boldsymbol{\alpha}_{M}) \otimes \mathbf{F}_{M}^{t}
\end{equation}

\noindent where $\otimes$ is the element-wise product. The intuition behind EAFG implies that the feature fusion module computes the weighted contribution of the PCA feature maps $\mathbf{F}_{M}^{p}$ and the fused features $\mathbf{F}^{f}_{M}$ is the weighted sum of the inputs. This improves the network's learning since PT-Fusion learns to adaptively compute the contribution of the features instead of a black box concatenation. Once the fused feature maps are computed, $\mathbf{F}^{f}_{M}$ is fed to the decoder for generating the network predictions.

Traditional decoders in U-Net architectures involve skip connections that concatenate the encoded features with the decoder outputs to preserve spatial information extracted by the encoders. However, since PT-Fusion comprises two encoding branches, concatenating the feature vectors from the two encoders with the decoder outputs tends to be memory intensive. Instead, PT-Fusion introduces AEDB to fuse the encoder features with the decoder output $\mathbf{D}_{m-1}$ of the layer $m-1$, analogous to the skip connection operation in the conventional U-Net. First, the features of the $m$ encoder layers are passed through EAFG to extract the fused features $\mathbf{F}_{m}^{f}$. Then, $\mathbf{F}_{m}^{f}$ and $\mathbf{D}_{m-1}$ are each passed through $1 \times 1$ convolution, $\mathbf{W}^{d}_{m}$ and $\mathbf{W}^{f}_{m}$ forming $\mathbf{D}'_{m}$ and $\mathbf{F}'_{m}$ as

\begin{align}
    \mathbf{D}'_{m} &= \mathbf{W}^{d}_{m}(\mathbf{D}_{m-1}), \\
    \mathbf{F}'_{m} &= \mathbf{W}^{f}_{m}(\mathbf{F}^{f}_{m}).
\end{align}

\noindent Accordingly, the attention map $\boldsymbol{\psi}_{m}$ is found by

\begin{equation}
    \boldsymbol{\psi}_{m} = \sigma \Biggl( \boldsymbol{W}_\psi \Bigl( \text{ReLU} ( \mathbf{D}'_{m} + \mathbf{F}'_{m}) \Bigl) \Biggl),
\end{equation}

\noindent where $\boldsymbol{W}_\psi$ is a learned $1 \times 1$ convolution. Finally, the output of the decoder $\mathbf{D}_{m}$ to the next decoder layer $m + 1$ is the element-wise product of between $\mathbf{D}_{m-1}$ and $\boldsymbol{\psi}_{m}$

\begin{equation}
    \mathbf{D}_{m} = \mathbf{D}_{m-1} \otimes \boldsymbol{\psi}_{m}.
\end{equation}

\noindent The EADB operations are repeated until the last decoding layer generate the predicted class logits $p_{s}$ for segmentation and depth map prediction $y_{d}$. By EADB, the network dynamically amplifies the important information from the encoders and decoders in a single feature representation, enabling accurate defect segmentation along and depth estimation. Additionally, computational and memory overheads are reduced since no redundant concatenations are present between the encoder-decoders output.

\begin{table*}[ht]
\centering
\caption{The ranges for PT-Fusion hyperparameter optimization along with its corresponding optimal values.}
\resizebox{0.9\textwidth}{!}{%
\resizebox{\columnwidth}{!}{%
\begin{tabular}{|c|c|c|}
\hline
\textbf{Hyperarameter}                       & \textbf{Hyperarameter Range}                & \textbf{Optimal Value}      \\ \hline
Number of Convolutional Layers  & {[}2, 3, 4, 5, 6, 7, 8{]}               & 5                           \\ \hline
Number of Convolutional Filters & {[}32, 64, 128, 256, 512, 1024, 2056{]} & {[}64, 128, 256, 512, 1024{]} \\ \hline
Convolutional Filters Kernel Size                & {[}3, 5, 7{]}               & 3                         \\
\hline
Dropout Coefficient             & {[}0.0, 0.5{]}                          & 0.0                         \\ \hline
Loss of Intermediate Layers     & Yes/No                                  & No                         \\ \hline
Batch Size                     & {[}8, 16, 32, 64, 128{]}                & 8                          \\ \hline
Number of Epochs                & {[}10, 25, 50, 75, 100, 500{]}               & 100                          \\ \hline
Learning Rate                & {[}$1e^{-3}$, $1e^{-4}$, $1e^{-5}${]}               & $1e^{-4}$                         \\
\hline
\end{tabular}%
}%
}
\label{table:model_params}
\end{table*}

\begin{algorithm}[t]
\caption{PT-Fusion Data Augmentation}
\begin{algorithmic}[1]
\Inputs{Original dataset \(\{\mathbf{S}_{i}\}\), $i = 1, \dots, 32$ \\ Number of Segments \(n = 100\) \\ Augmentation Factor \(\Gamma = 500\) \\ Additive Noise Variance \(\sigma^{2} = 0.005\)} \\
\vspace{0.2em}
\textbf{Outputs}: Augmented training and validation datasets

\State \textbf{Dataset Split}: 26 sequences for training, 6 for validation, 6 for testing
\For{each sequence \(\mathbf{S}_i\) in training and validation datasets}
    \State Divide \(\mathbf{S}_i\) into \(n = 100\) segments \(\{\bar{\mathbf{S}}_n\}_{n=1}^{100}\).
    \For{$t = 1$ to $\Lambda = 500$}
        \State Randomly sample \(I'_{r_n}\) from each segment:
        \[
        I'_{r_n} \sim \text{Uniform}(\min(\bar{\mathbf{S}}_{i,n}), \max(\bar{\mathbf{S}}_{i,n})).
        \]
        \State Form augmented sequence:
        \[
        \bar{\mathbf{S}}_{i,n}^{\prime} = \{I'_{r_1}, \dots, I'_{r_n}\}.
        \]
        \State Add Gaussian noise:
        \[
        \bar{\mathbf{S}}^{\text{noisy}}_{i, n} = \bar{\mathbf{S}}_{i,n}^{\prime} + \eta, \quad \eta \sim \mathcal{N}(0, \sigma^2).
        \]
        \State Compute PCA and TSR for \(\bar{\mathbf{S}}^{\text{noisy}}_{i, n}\)
    \EndFor
\EndFor
\end{algorithmic}
\label{alg:data_augmentation}
\end{algorithm}

\subsection{Data Augmentation} \label{subsection:augmentation}
As the IRT-PVC dataset comprises 38 inspection sequences, the dataset size is scarce and training the deep architecture of PT-Fusion remains challenging. While traditional augmentation techniques, such as random rotations, translations, and shearing, introduce spatial variability, they fall short in addressing the spatiotemporal nature of thermographic data. These sequences require augmentations that account for both spatial and temporal dimensions to effectively enhance dataset diversity. Therefore, this work proposes a spatiotemporal augmentation strategy, illustrated in Algorithm \ref{alg:data_augmentation}, to address this limitation. The proposed method involves three steps. First, subsets of thermographic sequences are randomly sampled from the original inspection sequences to introduce temporal variability. Second, additive Gaussian noise is applied on the randomly sampled sequences to further augment the dataset temporally. Finally, PCA and TSR tensors are generated from the randomly sampled sequences as new training samples in the expanded dataset. During training, traditional spatial augmentation techniques are applied dynamically on the generated tensors to enhance generalizibility of the network.

Before augmenting the dataset, 26 sequences out of the 38 in $\{\mathbf{S}_{i}\}$ are randomly assigned for training and 6 others for validation. The training and validation datasets are augmented using the proposed augmentation technique in Algorithm \ref{alg:data_augmentation}, while the rest of the dataset dedicated for testing remains unaugmented.

For each sequence in the training and validation datasets $\mathbf{S}_{i}$, where $i = 1, 2, \dots, 32$, $\mathbf{S}_{i}$ is divided into $n=100$ equally spaced segments $\bar{\mathbf{S}}_{n}$. From each segment, frames $I'_{r_{n}}$ are randomly sampled generating an augmented sequence $\bar{\mathbf{S}}'_{i,n}$ as

\begin{equation}
    \small
    \bar{\mathbf{S}}_{i,n}^{\prime} = \{ I'_{r_1}, \dots, I'_{r_{n}} \mid r_n \sim \text{Uniform}\bigl(\min(\bar{\mathbf{S}}_{i,n}), \max(\bar{\mathbf{S}}_{i,n})\bigr) \}
\end{equation}

\noindent where \( r_n \) is a randomly sampled index from segment $\bar{\mathbf{S}}_{i,n}$, drawn from the segment frame indices. This sequence generation process is repeated with an augmentation factor, $\Lambda = 500$, for each sequence. As a result, the dataset currently comprises 16000 samples instead of the original 32. To further expand the diversity of the dataset, additive zero-mean Gaussian noise \( \mathcal{N}(0, \sigma^2) \) is added to the augmented sequences by

\begin{equation}
    \bar{\mathbf{S}}^{\text{noisy}}_{i, n} = \bar{\mathbf{S}}_{i,n}^{\prime} + \eta, \quad \eta \sim \mathcal{N}(0, \sigma^2),
\end{equation}

\noindent where \( \sigma^2 = 0.05 \) controls the noise contribution to the sequence mimicking the noise characteristics in thermal imagers. From the augmented sequence $\bar{\mathbf{S}}^{\text{noisy}}_{i, n}$, PCA and TSR representations are generated, and traditional augmentation techniques such as random rotation, translation, flipping, and shearing are dynamically applied during training to grant PT-Fusion enhanced generalization.

\subsection{Implementation and Training Details} \label{subsection:training}
After formulating the PT-Fusion architecture incorporating the CNN heads, EAFG, and AEDB with its training data, a Bayesian optimization process over the network is employed to select its hyperparameters. The hyperparameters include the number of convolutional layers, number of filters, kernel sizes, and batch size. During the optimization process, the network weights are initialized using the built-in Glorot Uniform initializer in Pytorch. The resulting optimal network parameters are listed in Table \ref{table:model_params}. As outlined in Table \ref{table:model_params}, $m = 5$ convolutional layers with $3 \times 3$ kernels are incorporated for the PCA and TSR encoders. This also implies that the 5 AEDBs are employed to fuse the encoded features with the decoders at each $m$ layer.

We train two variants of the network that differ in the final prediction layer. The first predicts a multi-class segmentation mask, segmenting the pixels to defect classes defined by the defect's depth. For this variant, the training loss function is the cross-entropy loss $\mathcal{L}_{\beta}$ computed by

\begin{equation}
    \mathcal{L}_{\beta} = - \frac{1}{N} \sum_{i=1}^N \log\Bigl(\frac{\exp(x_{i,t})}{\exp(p_{i,s})}\Bigl),
\end{equation}

\noindent where $N$ is the number of pixels in the image, $x_{t}$ is the ground truth, and $p_{s}$ is the predicted class logits. The second variant of the network comprises of two outputs, 1) the binary segmentation mask $p_{b}$ and 2) the depth map prediction $y_{d}$. Hence, the loss function $\mathcal{L}_{\gamma}$ for training the network is a combination of binary cross entropy loss, $\mathcal{L}_{BCE}$, for segmentation and L1 loss, $\mathcal{L}_{1}$, for depth estimation as

\begin{equation}
    \mathcal{L}_{\gamma} = \mathcal{L}_{BCE} + \lambda \mathcal{L}_{1},
\end{equation}

\noindent where

\begin{equation}
    \mathcal{L}_{BCE} = - \frac{1}{N} \sum_{i=1}^N \big[ x_{i,t} \log(p_{i,b}) + (1 - x_{i,t}) \log(1 - p_{i,b}) \big],
\end{equation}

\begin{equation}
    \mathcal{L}_{L1} = \frac{1}{N} \sum_{i=1}^N \lvert y_i - y_{i,d} \rvert ,
\end{equation}

\noindent $\lambda = 0.5$ is a weighing factor, and $y_{i}$ is the depth map ground truth. The two variants of the network are trained for a two-fold validation routine discussed in section \ref{section:exp}. It is worth highlighting that both networks are trained with a batch size of 8 using an ADAM optimizer with a learning rate of $1e^{-4}$, as depicted in Table \ref{table:model_params}.

\begin{table*}[p]
\centering
\renewcommand{\arraystretch}{2}
\caption{Qualitative comparisons between state-of-the-art networks for multi-class defect segmentation, U-Net, attention U-Net, and 3D-CNN, against PT-Fusion.}
\resizebox{\textwidth}{!}{%
\begin{tabular}{|>{\centering\arraybackslash}m{1.25cm}|>{\centering\arraybackslash}m{1.5cm}|>
{\centering\arraybackslash}m{3cm}|>
{\centering\arraybackslash}m{3cm}|>{\centering\arraybackslash}m{3cm}|}
\hline
\textbf{Method} & \textbf{Input} & \textbf{Sample \#1} & \textbf{Sample \#2} & \textbf{Sample \#3} \\ \hline
\textbf{Attention U-Net} & PCA & {\vspace{3pt}\includegraphics[width=2.75cm]{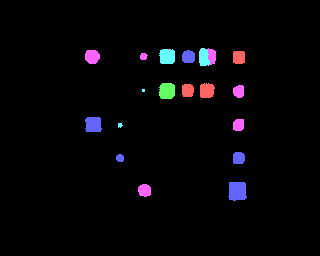}} & {\vspace{3pt}\includegraphics[width=2.75cm]{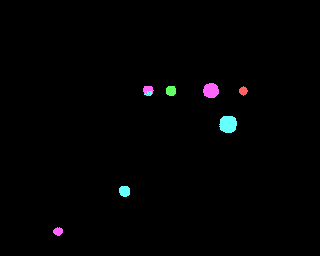}} & {\vspace{3pt}\includegraphics[width=2.75cm]{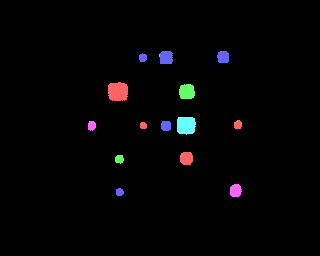}} \\ \hline
\textbf{Attention U-Net} & TSR & {\vspace{3pt}\includegraphics[width=2.75cm]{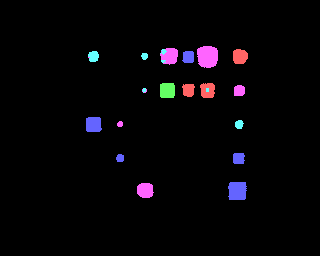}} & {\vspace{3pt}\includegraphics[width=2.75cm]{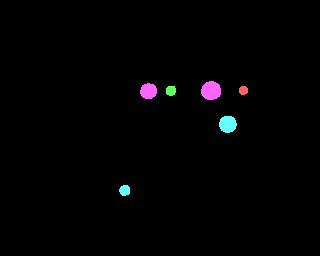}} & {\vspace{3pt}\includegraphics[width=2.75cm]{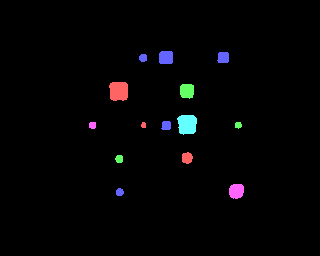}}\\ \hline
\textbf{Attention U-Net} & \makecell{PCA-TSR \\ Concatenated} & {\vspace{3pt}\includegraphics[width=2.75cm]{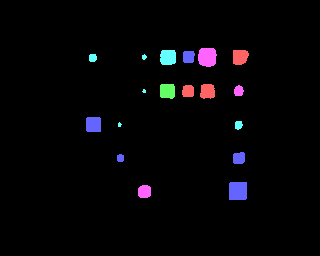}} & {\vspace{3pt}\includegraphics[width=2.75cm]{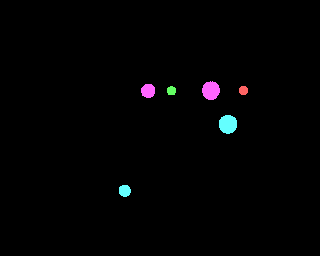}} & {\vspace{3pt}\includegraphics[width=2.75cm]{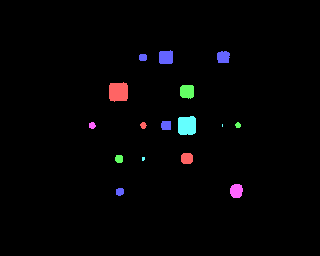}} \\ \hline
\textbf{U-Net} & \makecell{PCA-TSR \\ Concatenated} & {\vspace{3pt}\includegraphics[width=2.75cm]{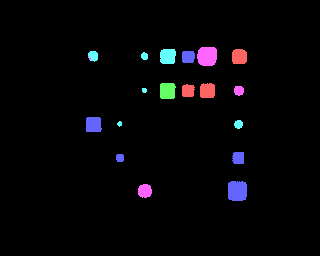}} & {\vspace{3pt}\includegraphics[width=2.75cm]{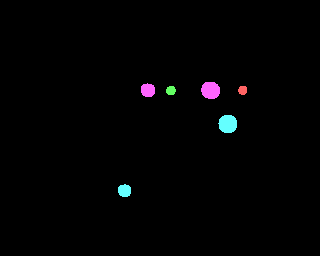}} & {\vspace{3pt}\includegraphics[width=2.75cm]{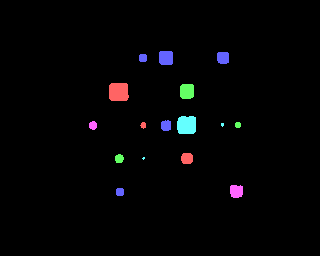}} \\ \hline
\textbf{3D-CNN} & Sequence & {\vspace{3pt}\includegraphics[width=2.75cm]{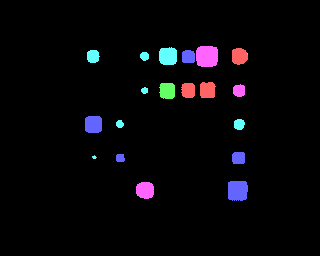}} & {\vspace{3pt}\includegraphics[width=2.75cm]{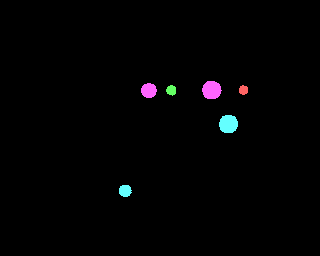}}  & {\vspace{3pt}\includegraphics[width=2.75cm]{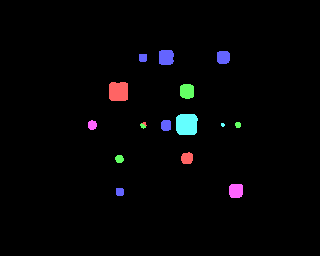}}  \\ \hline
\textbf{Ours} & \makecell{PCA-TSR \\ Fused} & {\vspace{3pt}\includegraphics[width=2.75cm]{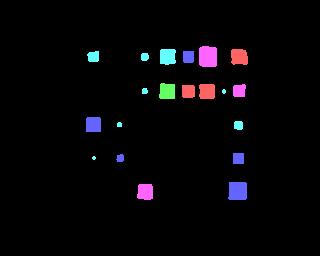}} & {\vspace{3pt}\includegraphics[width=2.75cm]{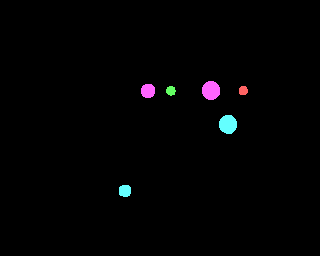}} & {\vspace{3pt}\includegraphics[width=2.75cm]{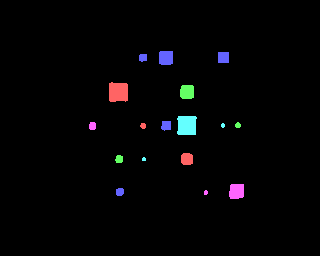}} \\ \hline
\textbf{GT} & N/A & {\vspace{3pt}\includegraphics[width=2.75cm]{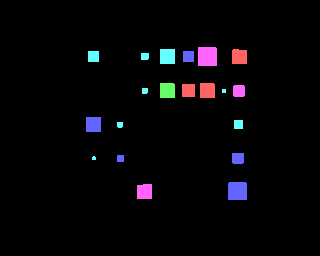}}  & {\vspace{3pt}\includegraphics[width=2.75cm]{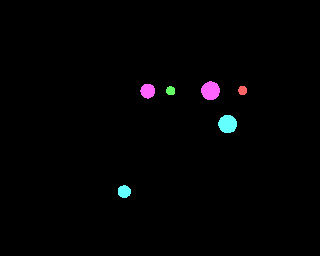}}  & {\vspace{3pt}\includegraphics[width=2.75cm]{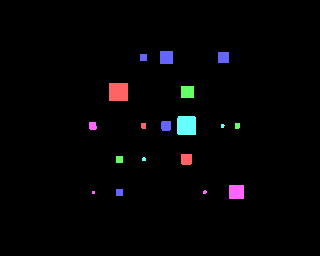}}  \\ \hline
\end{tabular}
}
\label{table:qual_comparison}
\end{table*}

\section{Experiments} \label{section:exp}
\subsection{Experimental Validation}
Given the training, validation, and testing splits, the two variants of PT-Fusion are tested on the unaugmented testing dataset in two corresponding testing routines. The first variant is trained to segment defects to different depth classes; thereby, multi-class segmentation, see section \ref{subsection:seg_eval}. For such type of problem, the common evaluation metrics, mean Intersection over Union mIoU, Recall, and Precision, are employed for evaluation and are defined by

\begin{equation}
    \text{mIoU} = \frac{1}{C} \sum_{c=1}^C \frac{|P_c \cap G_c|}{|P_c \cup G_c|}
\end{equation}

\begin{equation}
    \text{Recall} = \frac{1}{C} \sum_{c=1}^C \frac{\text{TP}_c}{\text{TP}_c + \text{FN}_c}
\end{equation}

\begin{equation}
    \text{Precision} = \frac{1}{C} \sum_{c=1}^C \frac{\text{TP}_c}{\text{TP}_c + \text{FP}_c},
\end{equation}

\noindent where $P_{c}$ and $G_{c}$ are the predicted and ground truth sets of pixels for class $c$, respectively. Similarly, TP${}_{c}$, FP${}_{c}$, and FN${}_{c}$ are the predicted true positives, false positives, and false negatives, respectively.

The second variant of the network is trained for simultaneous binary segmentation and depth estimation. Thus, the evaluation metrics are the IoU and the mean absolute error for depth defined as

\begin{equation}
    \text{IoU} = \frac{|P \cap G|}{|P \cup G|}
\end{equation}

\begin{equation}
    \text{MAE} = \frac{1}{N} \sum_{i=1}^N |y_g - y_d|,
\end{equation}

\noindent where $P$ and $G$ are the predicted and ground truth segmentation masks. This evaluation is discussed in section 
\ref{subsection:depth_eval}. It is worth mentioning that the two testing routines also present benchmarks of PT-Fusion against state-of-the-art defect segmentation and depth estimation models. Finally, the effect of the augmentation technique is discussed in section \ref{subsection:augm_eval} to demonstrate its need to train the deep architecture of PT-Fusion.

\subsection{Defect Segmentation Evaluation} \label{subsection:seg_eval}
The straightforward approach for defect segmentation using multi-modal representations is by concatenating the PCA and TSR images at the input. However, we show that such black box implementation does not provide optimal learning performance for the network. In general, it is always beneficial to guide the network architecture to inherently learn the contributions of the representations and dynamically fuse their features. Table \ref{table:qual_comparison} demonstrates qualitative comparisons of the predicted segmentation masks by a black box U-Net \cite{pt_dataset}, attention U-Net \cite{attention_unet}, 3D-CNN \cite{3d_cnn}, and PT-Fusion. The visualized results are also quantified in Table \ref{table:seg_metrics} in terms of the mIoU, recall, and precision, where PT-Fusion is also benchmarked against state-of-the-art defect segmentation networks.

\begin{table}[h]
\centering
\caption{PT-Fusion validated and benchmarked against state-of-the-art defect segmentation network architectures, Attention U-Net \cite{attention_unet} and 3D-CNN \cite{3d_cnn} in terms of mIoU, Recall, and Precision.}
\label{table:seg_metrics}
\resizebox{0.49\textwidth}{!}{%
\begin{tabular}{|c|c|c|c|c|}
\hline
\textbf{Method} &
  \textbf{Input} &
  \textbf{mIoU} &
  \textbf{Recall} &
  \textbf{Precision} \\ \hline
\textbf{\begin{tabular}[c]{@{}c@{}}Attention\\ U-Net\end{tabular}} &
  PCA &
  0.721 &
  0.696 &
  0.722 \\ \hline
\textbf{\begin{tabular}[c]{@{}c@{}}Attention\\ U-Net\end{tabular}} &
  TSR &
  0.711 &
  0.708 &
  0.723 \\ \hline
\textbf{\begin{tabular}[c]{@{}c@{}}Attention\\ U-Net\end{tabular}} &
  \begin{tabular}[c]{@{}c@{}}PCA-TSR\\ Concatenated\end{tabular} &
  0.804 &
  0.827 &
  0.864 \\ \hline
\textbf{U-Net} &
  \begin{tabular}[c]{@{}c@{}}PCA-TSR\\ Concatenated\end{tabular} &
  0.766 &
  0.801 &
  0.813 \\ \hline
\textbf{3D-CNN} &
  Sequence &
  0.784 &
  0.821 &
  0.847 \\ \hline
\textbf{Ours} &
  \textbf{\begin{tabular}[c]{@{}c@{}}Fused\\ PCA-TSR\end{tabular}} &
  \textbf{0.869} &
  \textbf{0.907} &
  \textbf{0.939} \\ \hline
\end{tabular}%
}
\end{table}

The obtained results outline several key findings. First, model performances fall around $0.7$ when trained on either PCA and TSR representations. When these representations are concatenated at the input, the mIoU increases to $0.766$ for U-Net and $0.804$ for attention U-Net. In addition, a $10\%$ increase in the recall and precision is witnessed. This demonstrates that the features from both representations are indeed complementary. However,concatenation at the input does not ensure optimal model performance. Hence, PT-Fusion architecture incorporating the fusion modules, EAFG and AEDB, effectively guides the network learning performance towards enhanced defect segmentation, where it outperforms the aforementioned methods by achieving an mIoU of $0.869$, a Recall of $0.907$, and a Precision of $0.939$. The proposed architecture inherently learns to adaptively identify the weighted contribution of the features to achieve state-of-the-art segmentation accuracy. Other methods, such as 3D-CNN, leverage 3D convolutions to extract spatial and temporal information simultaneously. Still, the network performance falls short compared to PT-Fusion. We hypothesize that this is due to the engineered sequence compression of the input discussed in \cite{3d_cnn}. In contrast, PT-Fusion leverages the enhanced defect visibility from PCA images and the temporal information of the pixel responses modeled in the TSR representation.

\begin{figure*}[!h]
    \centering
    \includegraphics[keepaspectratio=true,scale=0.475, width=\linewidth]{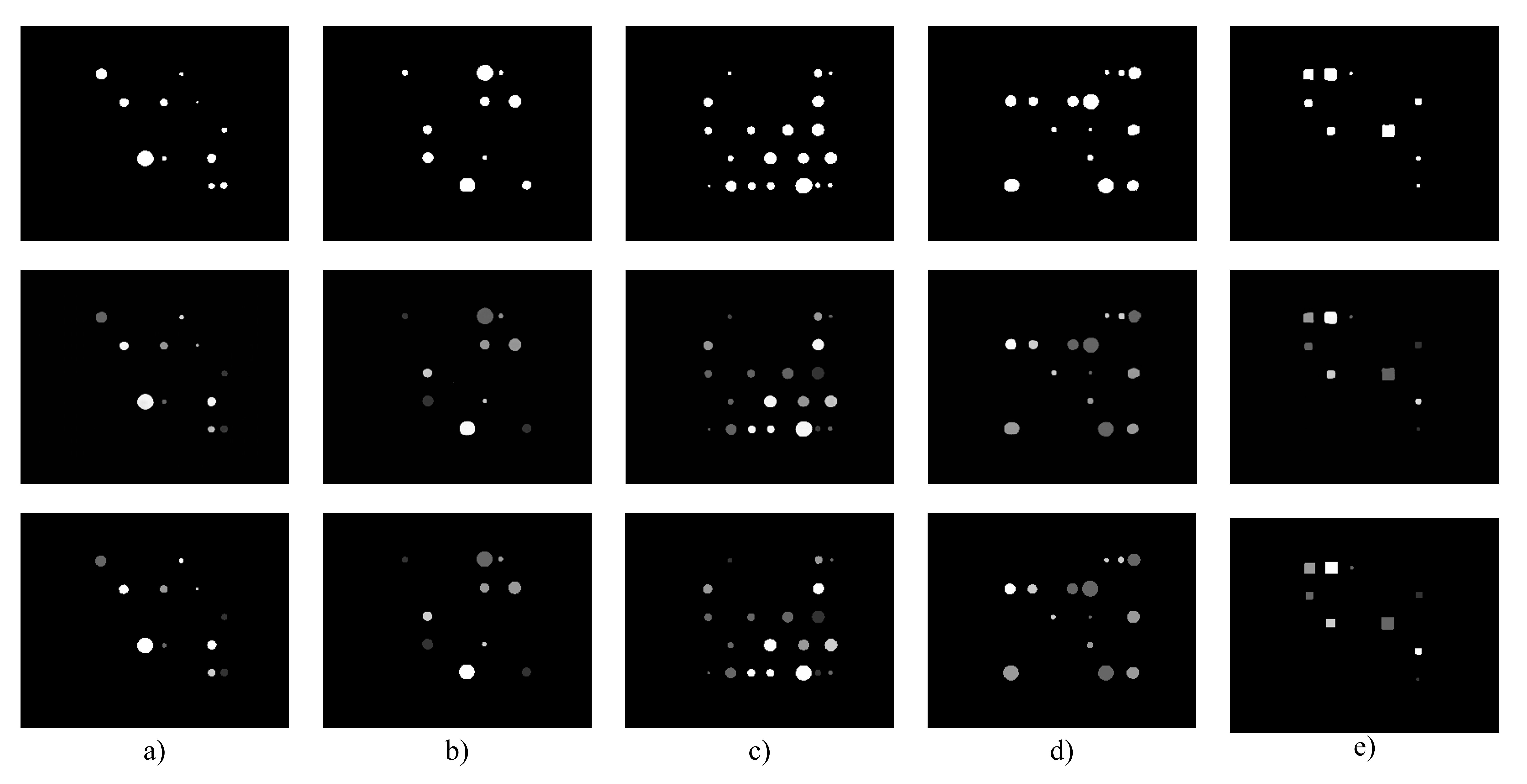}
    \caption{Simultaneous segmentation and depth estimation predictions by PT-Fusion for 5 samples a)-e). The segmentation and depth predictions, illustrated in the first and second rows, respectively, are compared against the ground truth in the third row.}
    \label{fig:seg_depth_imgs}
\end{figure*}

\subsection{Depth Estimation Performance} \label{subsection:depth_eval}
We also evaluate PT-Fusion architecture for depth estimation. While the previous evaluation routine involved multi-class segmentation, binary segmentation, i.e. classification of pixels as defective or sound, and depth prediction is more intuitive in actual inspection scenarios. This motivated training in the second variant of the proposed network that classifies the input pixels as either defective or sound along with the defect depth. Fig. \ref{fig:seg_depth_imgs} shows the predicted binary segmentation masks and depth maps for three samples in the IRT-PVC dataset. Table \ref{table:seg_depth_metrics} presents the performance metrics, IoU and MAE, and benchmarks against U-Net, attention U-Net, and 3D-CNN similar to the defect segmentation evaluation discussed in section \ref{subsection:seg_eval}.

\begin{table}[h]
\centering
\caption{PT-Fusion validated and benchmarked against state-of-the-art architectures \cite{pt_dataset, attention_unet,3d_cnn} for simultaneous defect segmentation and depth estimation in terms of IoU and MAE.}
\label{table:seg_depth_metrics}
\resizebox{0.49\textwidth}{!}{%
\begin{tabular}{|c|c|c|c|}
\hline
\textbf{Method}          & \textbf{Input}         & \textbf{IoU}   & \textbf{MAE (cm)}    \\ \hline
\textbf{Attention U-Net} & PCA                    & 0.748          & 0.0120          \\ \hline
\textbf{Attention U-Net} & TSR                    & 0.787          & 0.0119          \\ \hline
\textbf{Attention U-Net} & \begin{tabular}[c]{@{}c@{}}PCA-TSR\\ Concatenated\end{tabular} & 0.839 & 0.0103 \\ \hline
\textbf{U-Net}           & \begin{tabular}[c]{@{}c@{}}PCA-TSR\\ Concatenated\end{tabular} & 0.828 & 0.0108 \\ \hline
\textbf{3D-CNN}          & Sequence               & 0.823          & 0.0113          \\ \hline
\textbf{Ours}            & \textbf{Fused PCA-TSR} & \textbf{0.882} & \textbf{0.0082} \\ \hline
\end{tabular}%
}
\end{table}

Similar findings to the defect segmentation evaluation can be drawn. The IoU for all networks shows an improvement compared to the mIoU in Table \ref{table:seg_metrics}. This is expected since binary segmentation tends to be a simplified version of multi-class segmentation. In addition, since the network predicts the segmentation mask and depth map simultaneously, the segmentation performance is improved as it is guided by the predicted depth of the defect. On the other hand, by comparing the individual performances of the networks, PT-Fusion outperforms the state-of-the-art network architectures by adaptively computing the contributions of multi-modal features. This is also evident in the depth estimation task since it heavily relies on the pixel's temporal responses along with the spatial features in the representations. Accordingly, PT-Fusion harnesses the temporal information embedded in TSR, in addition to the enhanced spatial information in the PCA representation leading to improved segmentation and depth estimation compared to U-Net, attention U-Net, and 3D-CNN.

\begin{figure}[t]
\center
\includegraphics[scale=0.29]{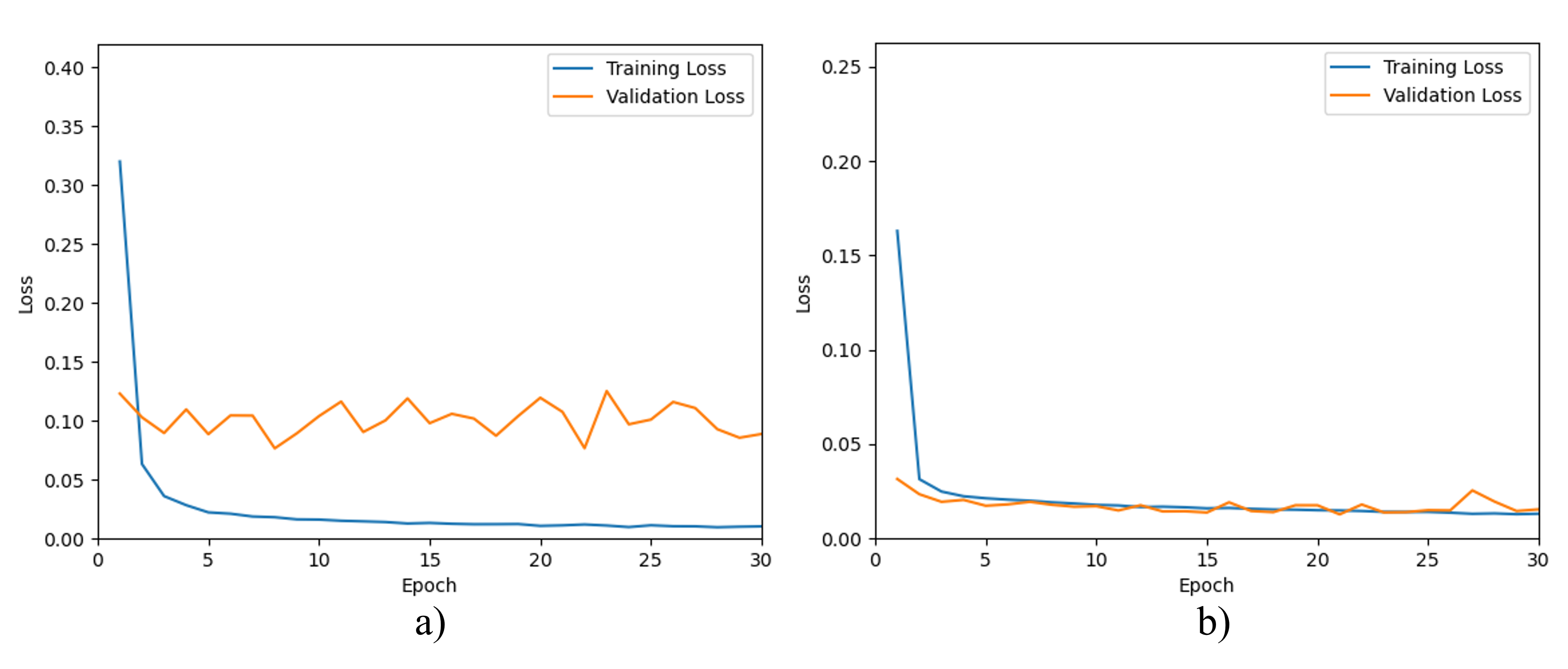}
\caption{Training loss for PT-Fusion trained on a) data augmented with common augmentation techniques and b) data augmented using the proposed augmentation strategy.}
\label{fig:augm_overfit}
\end{figure}

\begin{figure}[b]
\center
\includegraphics[scale=0.36]{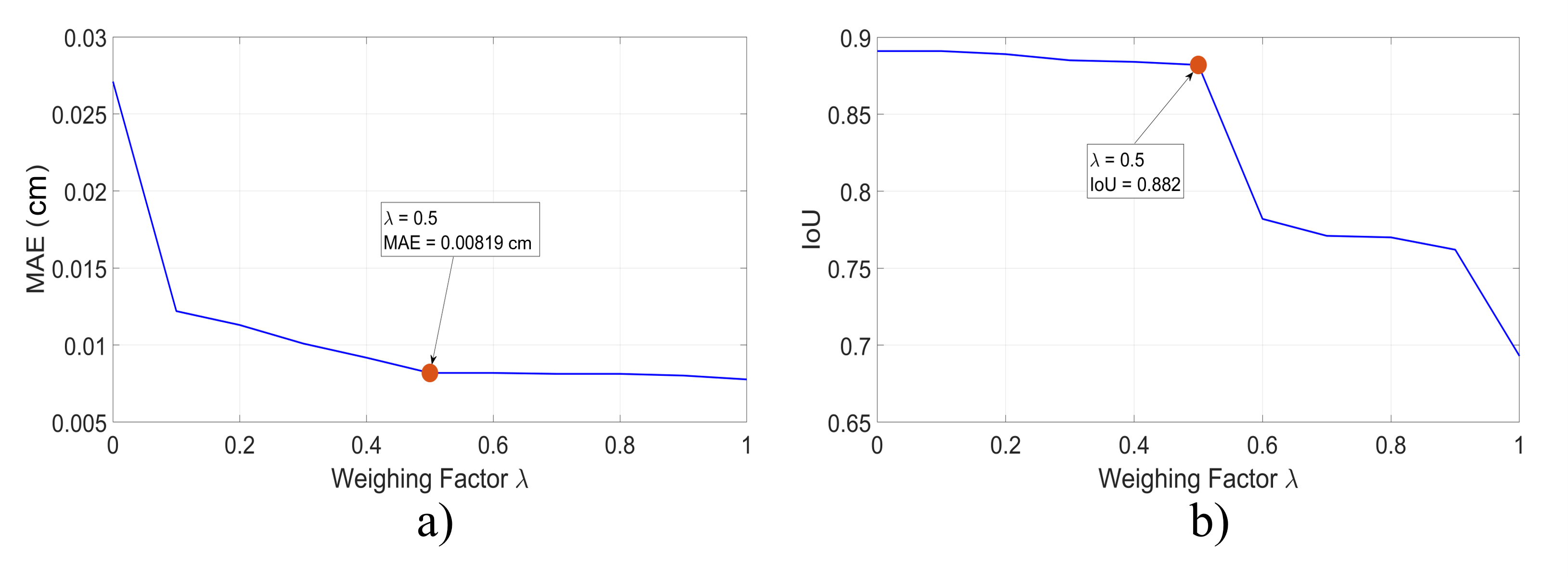}
\caption{Effect of the weighing factor $\lambda$ on IoU and MAE, where $\lambda = 0.5$ provides the optimal trade-off between both metrics and is utilized during training.}
\label{fig:ablation}
\end{figure}

\subsection{Effect of Data Augmentation}
\label{subsection:augm_eval}
The obtained network performance is heavily dependent on the diversity of the dataset. We demonstrate the need for the proposed augmentation strategy by training PT-Fusion on the original IRT-PVC dataset augmented using the common augmentation techniques, i.e., random translations, rotations, and shearing. Fig. \ref{fig:augm_overfit}a shows the training loss for PT-Fusion trained on the original dataset. As depicted in the figure, the validation loss tends to remain constant and does not decrease with the training loss. The training resulted in a corresponding testing IoU of $0.567$. This implies that the traditional augmentation techniques are still not sufficient to train the deep architecture of PT-Fusion.

In contrast, Fig. \ref{fig:augm_overfit}b shows the training loss for PT-Fusion trained on the IRT-PVC sequences augmented using the introduced augmentation approach along with the aforementioned common augmentation techniques. The devised data augmentation strategy proves to be sufficient to grant PT-Fusion the required generalization and segmentation performance, where the obtained testing IoU is $0.882$. While traditional augmentation methods tend to virtually increase the sizes of the datasets, the augmentation is done spatially. In addition, PT-based inspection data analysis inherently involves spatiotemporal data requiring spatial and temporal augmentations simultaneously. Hence, our data augmentation method incorporates spatial and temporal augmentations, ensuring dataset diversity and improved training of the PT-Fusion network.

\subsection{Ablation Studies}
We perform two further studies on the network architecture trained for simultaneous segmentation and depth estimation: 1) The weighing factor $\lambda$ in the loss function $\mathcal{L}_{\gamma}$ and 2) incorporating self-attention mechanisms instead of the fusion modules, EAFG and AEDB, for harnessing long-range dependencies in the encoded features. The studies are based on the IoU metric for binary segmentation and MAE for depth estimation. With regards to $\lambda$, the MAE tends to decrease with increasing lambda till $\lambda=0.5$. Then, the MAE remains approximately consistent. In contrast, the IoU tends to remain approximately $0.88$ till $\lambda=0.5$ and starts to decrease dramatically. This is because the loss function importance shifts to the depth estimation task. Hence, based on Fig. \ref{fig:ablation}, one can infer that the optimal value of $\lambda$ is 0.5 balancing the accuracy trade-off between the MAE and IoU.

The second study involves incorporating self-attention and cross-attention mechanisms in combination with or in place of EAFG and AEDB fusion blocks. Self-attention and cross-attention mechanisms allow the model to capture long-range dependencies between the obtained feature maps from PT-Fusion encoders. Table \ref{table:ablation} compares PT-Fusion performance with and without the attention mechanisms in terms of IoU and MAE. First, it can be inferred that when combining attention mechanisms with AEDB and EAFG, no significant improvement in performance is witnessed in terms of IoU and depth MAE. On the other hand, by fully employing the attention mechanisms as fusion modules, improvements in the obtained performances are approximately $2\%$ with a 5-fold increase in the number of trainable parameters. This implies that the needed features for inspection of the IRT-PVC samples are mostly local. The reason behind this is that the PT setup involves the inspection of PVC samples, which are generally heat insulators. Hence, heat remains locally confined within the defects, with no global features in the input thermographic representations, and EAFG and AEDB are sufficient to capture these features for defect segmentation and depth estimation.

\begin{table}[h]
\centering
\caption{Performance Comparison between PT-Fusion architecture with self-attention cross-attention mechanisms in place of EAFG and AEDB fusion blocks.}
\label{table:ablation}
\resizebox{0.45\textwidth}{!}{%
\begin{tabular}{|c|c|c|}
\hline
\textbf{\begin{tabular}[c]{@{}c@{}}Fusion Modules\end{tabular}} & \textbf{IoU} & \textbf{MAE (cm)} \\ \hline
EAFG + AEDB                                                       & 0.882        & 0.0082            \\ \hline
Self-Attention +  Cross-Attention                                 & 0.901        & 0.0072            \\ \hline
EAFG + Cross-Attention & 0.883 & 0.0080 \\ \hline

Self-Attention + AEDB & 0.884 & 0.0084 \\ \hline

EAFG + AEDB + Self-Attention & 0.888 & 0.0079 \\ \hline
\end{tabular}%
}
\end{table}

\section{Conclusion} \label{section:conc}
Infrared thermography (IRT) became prominent in the field of NDT, enabling the detection of subsurface defects in industrial components. Particularly, pulse thermography (PT) is valued for its efficiency and fast nature. Current state-of-the-art PT inspection methods rely on PCA and TSR independently as image modalities to train defect segmentation and depth estimation learning-based PT inspection models. This limits the performance of the trained networks as these representations possess complementary semantic features. Thus, this work proposes PT-Fusion, a multi-modal attention network that fuses PCA and TSR for enhanced segmentation and depth estimation of subsurface defects. The network architecture introduces two fusion modules, namely, Encoder Attention Fusion Gate (EAFG) to adaptively learn the weighted contributions of the input representations and Attention Enhanced Decoding Block (AEDB) to dynamically fuse the encoded features with outputs of the PT-Fusion decoders. In addition, due to the scarcity of PT datasets, a novel spatiotemporal data augmentation technique is proposed to mitigate this challenge and facilitate training of PT-Fusion. The network is tested on the Université Laval IRT-PVC dataset for defect segmentation and depth estimation. PT-Fusion is also benchmarked against state-of-the-art PT network architectures, such as U-Net, attention U-Net, and 3D-CNN. The results demonstrate that PT-Fusion outperforms the aforementioned models with an IoU of $0.882$ and depth estimation error of $0.0082$ mm. Future research efforts will be directed towards extending PT-Fusion to lock-in thermography setups and inspection of other materials such as CFRP and PLA. In addition, experimental studies will be conducted on the efficacy of self and cross-attention mechanisms during the inspection of materials with varying thermal conductivities.

\bibliographystyle{IEEEtran}
\bibliography{main.bib}

\vskip -1\baselineskip plus -1fil

\begin{IEEEbiography}[{\includegraphics[width=1in,height=1.25in,clip,keepaspectratio]{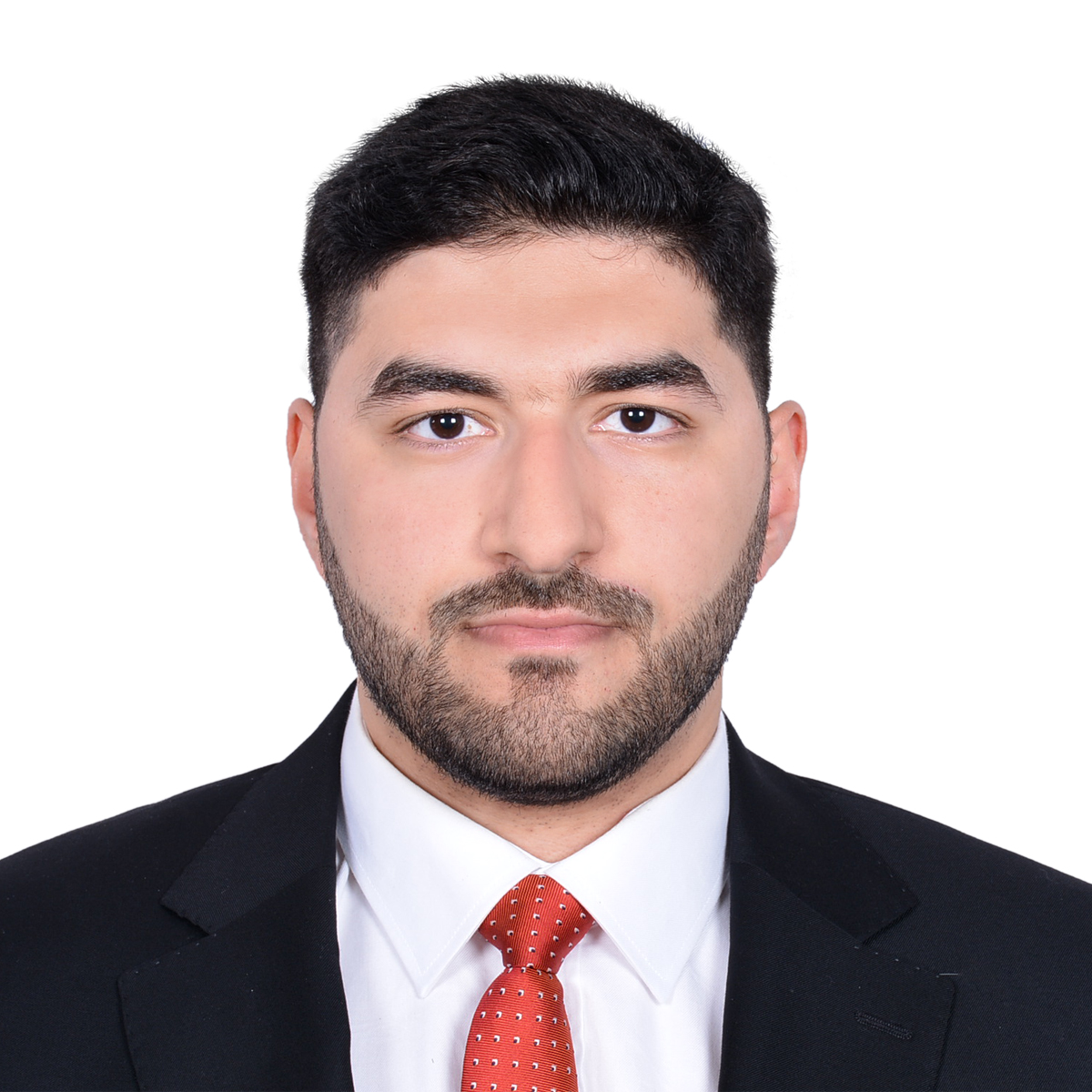}}]%
{Mohammed Salah} received his BSc. in Mechanical Engineering from the American University of Sharjah, UAE, in 2020 and his MSc. in Mechanical Engineering from Khalifa University, Abu Dhabi, UAE, in 2022. He is currently pursuing his Ph.D. in Robotics at Khalifa University, Abu Dhabi, UAE. His research interests include robotic automation, thermography, and non-destructive testing.
\end{IEEEbiography}

\begin{IEEEbiography}[{\includegraphics[width=1in,height=1.25in,clip,keepaspectratio]{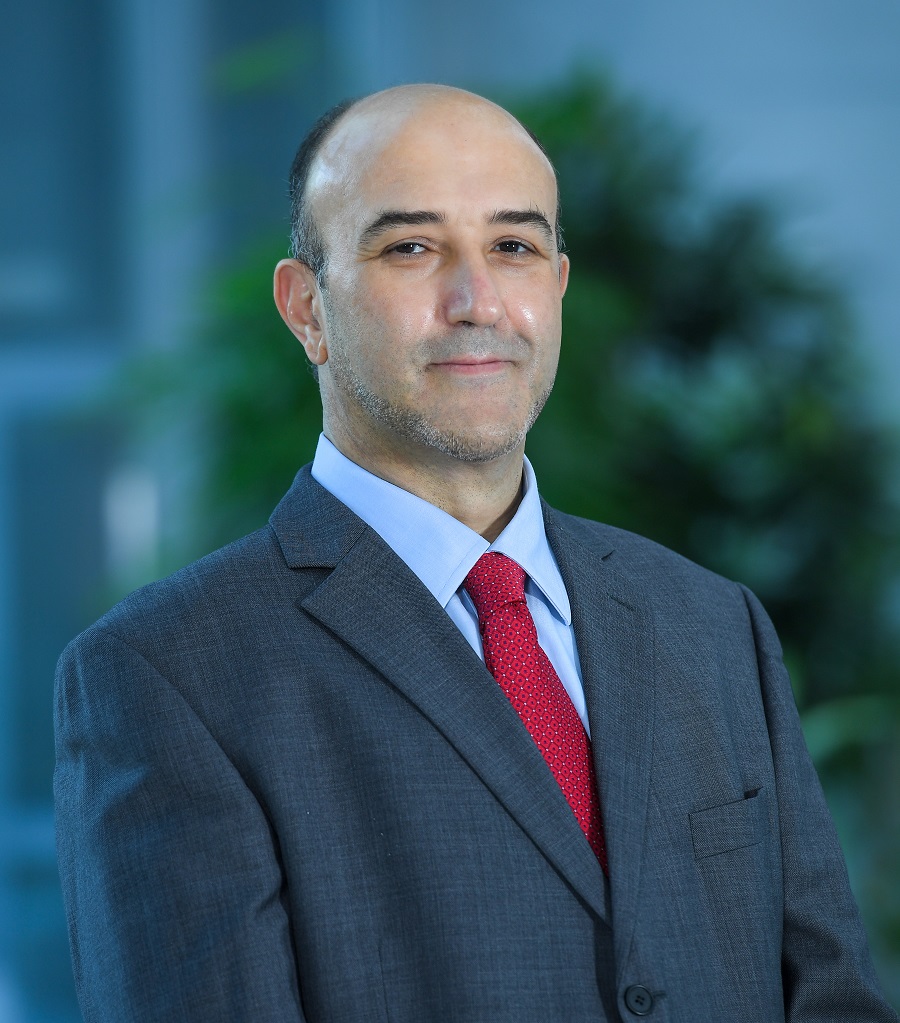}}] {Naoufel Werghi} received his Habilitation and Ph.D. in Computer Vision from the University of Strasbourg. Currently, he is a full professor in the computer science department at Khalifa University, Abu Dhabi, United Arab Emirates. He is the theme leader of the Artificial  Intelligence and Big Data Pipelines theme in the Cyber-Physical Security System Center (C2PS) at Khalifa University. His research interests span computer vision and machine learning, where he has been leading several funded projects related to biometrics, medical imaging, remote sensing, surveillance, and intelligent systems. 
\end{IEEEbiography}

\begin{IEEEbiography}[{\includegraphics[width=1in,height=1.25in,clip,keepaspectratio]{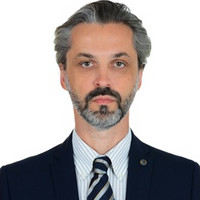}}]{Davor Svetinovic}
is a professor at the Department of Computer Science, Khalifa University, Abu Dhabi, UAE. He received his doctorate in computer science from the University of Waterloo, Waterloo, ON, Canada, in 2006. Previously, he worked at WU Wien, TU Wien, Austria, and Lero -- the Irish Software Engineering Center, Ireland. He was a visiting professor and a research affiliate at MIT and MIT Media Lab, MIT, USA. Davor has extensive experience working on complex multidisciplinary research projects. He has published over 100 papers in leading journals and conferences and is a highly cited researcher in blockchain technology. His research interests include cybersecurity, blockchain technology, crypto-economics, trust, and software engineering. His career has furthered his interest and expertise in developing advanced research capabilities and institutions in emerging economies. He is a Senior Member of IEEE and ACM (Lifetime), and an affiliate of the Mohammed Bin Rashid Academy of Scientists.
\end{IEEEbiography}

\begin{IEEEbiography}
   [{\includegraphics[width=1in]{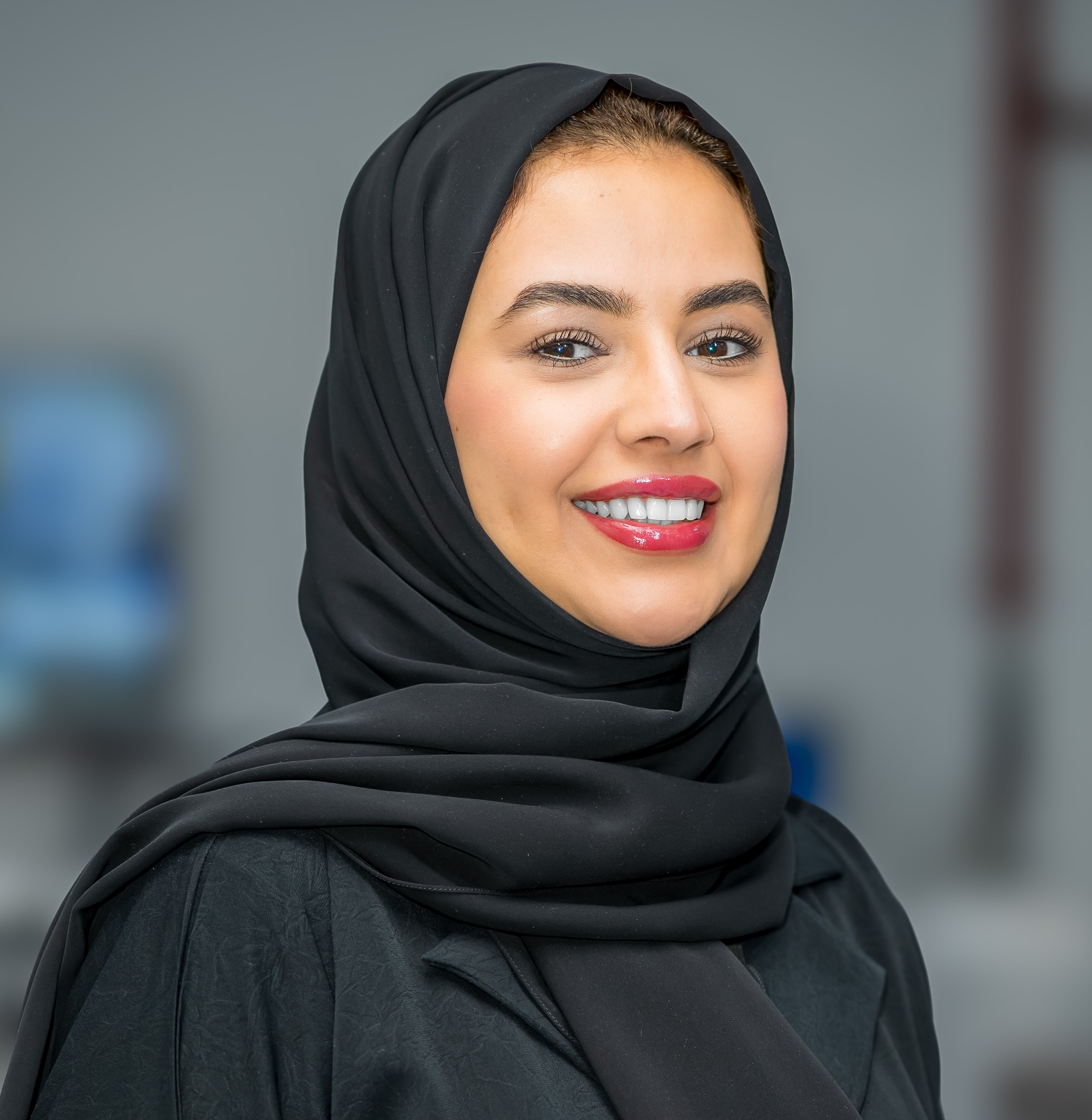}}]
{Yusra Abdulrahman} (IEEE member) received the B.Sc. degree from The University of
Arizona, in 2014, and the M.Sc. and Ph.D. degrees
from Massachusetts Institute of Technology and
Masdar Institute of Science and Technology
Cooperative Program (MIT and MICP), in
2016 and 2020, respectively. She is currently
an Assistant Professor at the Department of
Aerospace Engineering, Khalifa University of Science and Technology. Her
expertise lies in robotics, artificial intelligence
(AI), and non-destructive testing (NDT). 
\end{IEEEbiography}

\end{document}